\documentclass[runningheads]{llncs}
\usepackage{multirow}
\usepackage{comment}
\usepackage{enumitem}
\usepackage{mdframed}
\usepackage{tipa}
\usepackage{enumitem}
\usepackage{paralist,multirow,url}
\usepackage{amssymb}
\usepackage{pifont}
\usepackage{paralist,bbding,pifont}
\usepackage{amsmath,graphicx}
\usepackage{booktabs}
\usepackage{geometry}
\begin{document}
\title{I$^2$KD-SLU:~ An Intra-Inter Knowledge Distillation Framework for Zero-Shot Cross-Lingual Spoken Language Understanding}
\titlerunning{I$^2$KD-SLU}
%
\author{Tianjun Mao \and Chenghong Zhang\thanks{Corresponding author: Chenghong Zhang.}}
\authorrunning{T. Mao et al.}
%
\institute{School of Management, Fudan University, Shanghai, China 
\\
\email{tjmao22@m.fudan.edu.cn, chzhang@fudan.edu.cn}}
\maketitle              
\begin{abstract}
Spoken language understanding~(SLU) typically includes two subtasks: intent detection and slot filling. Currently, it has achieved great success in high-resource languages, but it still remains challenging in low-resource languages due to the scarcity of labeled training data. Hence, there is a growing interest in zero-shot cross-lingual SLU. Despite of the success of existing zero-shot cross-lingual SLU models, most of them neglect to achieve the mutual guidance between intent and slots. To address this issue, we propose an \textbf{I}ntra-\textbf{I}nter \textbf{K}nowledge \textbf{D}istillation framework for zero-shot cross-lingual \textbf{S}poken \textbf{L}anguage \textbf{U}nderstanding~(I$^2$KD-SLU) to model the mutual guidance. Specifically, we not only apply intra knowledge distillation between intent predictions or slot predictions of the same utterance in different languages, but also apply inter knowledge distillation between intent predictions and slot predictions of the same utterance. Our experimental results demonstrate that our proposed framework significantly improves the performance compared with the strong baselines and achieves the new state-of-the-art performance on the MultiATIS++ dataset, obtaining a significant improvement over the previous best model in overall accuracy.

\end{abstract}

\section{Introduction}

Spoken language understanding~(SLU) aims to extract the semantic components from user queries~\cite{tur2011spoken,huang2022mtl,chen2022leveraging,ijcai2022p0565,huang2020federated,zhou2021pin,cheng_F,huang2021sentiment,cheng_C}, which is an important component in the task-oriented dialogue systems. SLU typically involves two subtasks: intent detection and slot filling. Intent detection obtains the user's intent from the input utterance and slot filling recognizes entities carrying detailed information of the intent. Deep neural network techniques have achieved remarkable results in SLU, but they require extensive labeled training data, which limits their scalability to languages with little or no training data. To address this limitation, zero-shot cross-lingual SLU generalization has received attention, which uses labeled data from high-resource languages to transfer trained models to low-resource target languages.

As deep learning applied to various tasks~\cite{NEURIPS2020_dc49dfeb,cheng2023s,tian2021continuous,guo2020powering,tian2023designing,yao2023poserac}, machine translation technique is first introduced in data-based transfer methods to convert source utterances into targets~\cite{upadhyay2018almost,schuster2019cross,xu2020end}. Nevertheless, for some exceptionally low-resource languages, machine translation might be undependable or inaccessible~\cite{upadhyay2018almost}. To tackle this issue, some studies~\cite{qin2020cosda} aligned source languages with multiple target languages using bilingual dictionaries to randomly replace some words in the utterance with translation words in other languages, while others~\cite{qin2022gl,liang2022label} have applied contrastive learning to achieve explicit alignment and improve performance. However, previous works have neglected the mutual guidance between intent and slots. Normally, intents and slots are related. So it is beneficial to model the mutual guidance between intents and slots in zero-shot cross-lingual spoken language understanding.

In this paper, we propose an intra-inter knowledge distillation framework for zero-shot cross-lingual spoken language understanding termed I$^2$KD-SLU based on multilingual BERT (mBERT)~\cite{devlin2019bert}. mBERT is a pre-trained contextual model trained on a large corpus of multiple languages, and it has shown significant progress in achieving zero-shot cross-lingual SLU. Specifically, for intra knowledge distillation, we apply knowledge distillation between intent predictions or slot predictions of the same utterance in different languages. For inter knowledge distillation, we apply knowledge distillation between intent predictions and slot predictions of the same utterance. Intra knowledge distillation helps to transfer knowledge from different languages and inter knowledge distillation helps to achieve the mutual guidance between intents and slots. Experiment results on the public benchmark dataset MultiATIS++~\cite{xu2020end} demonstrate that I$^2$KD-SLU significantly outperforms the previous best cross-lingual SLU models and analysis further verifies the advantages of our method.

 In summary, the contributions of this work can be concluded as follows:
\begin{compactitem}
\item To the best of our knowledge, we are the first to achieve mutual guidance between intent and slots for zero-shot cross-lingual SLU.
\item We propose an intra-inter knowledge distillation framework, where intra knowledge distillation promotes the knowledge transfer and inter knowledge distillation models the mutual guidance.
\item Experiments show that our method achieves a new state-of-the-art performance, obtaining an improvement of 3.0$\%$ over the previous best model in terms of average overall accuracy of 9 languages.
\end{compactitem}
\section{Related Work}
\subsection{Spoken Language Understanding} Intent detection and slot filling tasks are two typical sub-tasks of SLU~\cite{cheng2023acl,zhu2023acl,zhu2023icassp}. While slot filling can be challenging as decisions must be made for each word or token, it is applied in interesting use cases, as noted by \cite{gunaratna2021using}. In the past, these two tasks are performed independently, but recent research has shown that jointly optimizing them can improve accuracy~\cite{zhang2016joint,qin2021co,qin2021gl,xing2022co}. Contextual language models have also improved language encoding capabilities for joint NLU models compared to traditional static word embedding approaches. Despite there are lots of remarkable results in SLU, they all require extensive labeled training data, which limits their scalability to languages with little or no training data. As a result, the concept of zero-shot cross-lingual SLU generalization has gained traction, where models are trained using labeled data from high-resource languages and then transferred to low-resource target languages without additional training data. Lately, there have been encouraging results achieved by cross-lingual contextualized embeddings such as mBERT~\cite{devlin2019bert}. Several studies have concentrated on enhancing mBERT~\cite{xu2020end,qin2022gl,liang2022label,chen2019bert,razumovskaia2022crossing,ijcai2020p0533}. However, they both neglect to achieve the mutual guidance between intent and slots. In our work, we tackle this issue by applying intra and inter knowledge distillation.
\subsection{Knowledge Distillation}
Knowledge Distillation is a technique first proposed by \cite{hinton2015distilling}. The goal of knowledge distillation is to transfer the knowledge from a large, complex model which is known as the teacher to a smaller, simpler model which is known as the student. This is achieved by training the student model to mimic the behavior of the teacher model, using either the predicted outputs or intermediate representations of the teacher model. Existing knowledge distillation methods generally fall into two categories. The first category focuses on using dark knowledge~\cite{romero2014fitnets,you2017learning} and the second category focuses on sharing information about the relationships between the layers of the teacher model~\cite{yim2017gift,tarvainen2017mean}. In our method, we apply knowledge distillation to facilitate knowledge transfer between different languages and achieve the mutual guidance between intent and slots.

\section{Method}
\begin{figure*}[t]
  \centering
\includegraphics[width=.9\linewidth]{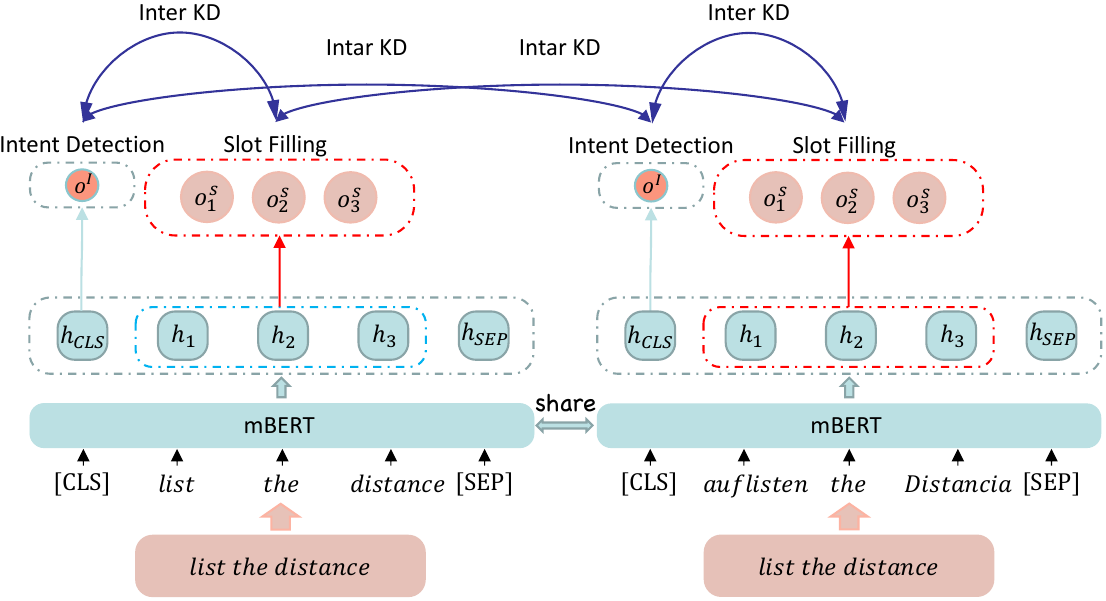}
  \caption{The overview of our I$^2$KD-SLU. Two models with the same architecture are trained on the original utterance and code-switched utterance, respectively. Intra knowledge distillation helps to transfer knowledge between different languages. Inter knowledge distillation helps to achieve mutual guidance between intent and slots.}
  \label{fig:method}
\end{figure*}

In this section, we first describe the background~($\S\ref{DEF}$) of zero-shot cross-lingual SLU. Then we introduce the main architecture of our framework I$^2$KD-SLU. Finally we introduce the final training objective~($\S\ref{obj}$). The overview of our framework is illustrated in Figure \ref{fig:method}.

\subsection{Background}
\label{DEF}
Intent detection and slot filling are two subtasks of SLU. Given an input utterance $\textbf{x} = (x_{1}, x_{2}, \dots, x_{n})$, where $n$ is the length of $\textbf{x}$. Intent detection is a classification task which predicts the intent $\boldsymbol{o}^I$. Slot filling is a sequence labeling task which maps each utterance $\textbf{x}$ into a slot sequence $\boldsymbol{o}^S=\left(o_{1}^{S}, o_{2}^{S}, \dots, o_{n}^{S}\right)$. Training a single model that can handle both tasks of intent detection and slot filling is a common practice as they are closely interconnected. Following previous work~\cite{goo2018slot}, the formalism is formulated as follows:
\begin{equation}
    (\boldsymbol{o}^{I}, \boldsymbol{o}^{S})=f(\textbf{x})
\end{equation}
where $f$ is the trained model.

The zero-shot cross-lingual SLU task involves training an SLU model in a source language and then adapting it directly to target languages without additional training. Specifically, given each instance $\mathbf{x}_{tgt}$ in the target language, the predicted intent and slot can be directly obtained by the SLU model $f$ which is trained on the source language:
\begin{equation}
    \left(\boldsymbol{o}_{tgt}^{I}, \boldsymbol{o}_{tgt}^{S}\right)=f\left(\mathbf{x}_{tgt}\right)
\end{equation}
where $tgt$ denotes the target language.
\subsection{Main Architecture}
\label{SLU}
Inspired by the success of pre-trained models~\cite{hou2020dynabert,dong2023polyvoice,cheng_g,huang2021audio,li2023generating,li2023unify,huang2021ghostbert,zhu_interspeech,li2023iccv,cheng2023m}, we obtain the representation $\mathbf{H}$ of the utterance $\textbf{x}$ by using mBERT~\cite{devlin2019bert} model:
\begin{equation}
    \mathbf{H} =  (\boldsymbol{h}_{\texttt{CLS}}, \boldsymbol{h}_{1}, \dots , \boldsymbol{h}_{n}, \boldsymbol{h}_{\texttt{SEP}})
\end{equation}
 where $\texttt{[CLS]}$ denotes the special symbol for representing the whole sequence, and $\texttt{[SEP]}$ can be used for separating non-consecutive token sequences. 

For intent detection, we input the utterance representation $\boldsymbol{h}_{\texttt{CLS}}$ to a classification layer to obtain the predicted intent:
\begin{equation}
\boldsymbol{o}^I=\operatorname{softmax}\left(\boldsymbol{W}^I \boldsymbol{h}_{\texttt{CLS}}+\boldsymbol{b}^{I}\right)
\end{equation}
where $\boldsymbol{W}^I$ and $\boldsymbol{b}^I$ denote the trainable matrices.

For slot filling, we follow \cite{wang2019cross} to utilize the representation of the first sub-token as the whole word representation and utilize the hidden state to predict each slot in the utterance:
 \begin{equation}
\boldsymbol{o}_t^{S}=\operatorname{softmax}\left(\boldsymbol{W}^{s} \boldsymbol{h}_{t}+\boldsymbol{b}^{s}\right)
 \end{equation}
 where $\boldsymbol{h}_{t}$ denotes the representation of the first sub-token of word $x_t$, $\boldsymbol{W}^s$ and $\boldsymbol{b}^s$ denote the trainable matrices.

We employ the code-switching approach~\cite{ijcai2020p0533} to leverage bilingual dictionaries~\cite{lample2018word} in generating multi-lingual code-switched utterance $\textbf{x}'$. We denote the intent prediction and slot prediction of the original utterance $\mathbf{x}$ as $\mathbf{o}^{Io}$ and $\mathbf{o}^{So}$, respectively. Similarly, we denote the intent prediction and slot prediction of the code-switched utterance $\mathbf{x}'$ as $\mathbf{o}^{Ic}$ and $\mathbf{o}^{Sc}$, respectively. Note that $\mathbf{o}^{So}$ and $\mathbf{o}^{Sc}$ both consist of all slot predictions in the corresponding utterance. 

We apply intra knowledge distillation to promote knowledge transfer between different languages, which includes two components. One is the Jensen-Shannon Divergence~(JSD) between the intent prediction of the original utterance and the intent prediction of the code-switched utterance. The other is the JSD between the slot prediction of the original utterance and the slot prediction of the code-switched utterance. The final intra knowledge distillation loss $\mathcal{L}_{intra}$ is as follows:
\begin{equation}
    \mathcal{L}_{intra} = \operatorname{JSD}(\mathbf{o}^{Io},\mathbf{o}^{Ic}) + \operatorname{JSD}(\mathbf{o}^{So},\mathbf{o}^{Sc})
\end{equation}

We also apply inter knowledge distillation to achieve mutual guidance between intent and slots, which also includes two components. One is the JSD between the intent prediction of the original utterance and the slot prediction of the original utterance. The other is the JSD between the intent prediction of the code-switched utterance and the slot prediction of the code-switched utterance. The final inter knowledge distillation loss $\mathcal{L}_{inter}$ is computed as follows:
\begin{equation}
    \mathcal{L}_{inter} = \operatorname{JSD}(\mathbf{o}^{Io},\operatorname{Avg}(\mathbf{o}^{So})) + \operatorname{JSD}(\mathbf{o}^{Ic},\operatorname{Avg}(\mathbf{o}^{Sc}))
\end{equation}
where $\operatorname{Avg}$ denotes averaging all slots prediction in the utterance.

\subsection{Training Objective}
\label{obj}
Following previous work~\cite{goo2018slot}, the intent detection objective $\mathcal{L}_I$ and the slot filling objective $\mathcal{L}_S$ are formulated as follows:
\begin{gather}
\mathcal { L } _ { I } \triangleq -\sum _ { i = 1 } ^ { n_{I} }  \hat { {\bf{y}} } _ { i} ^ {I } \log \left( {\bf{o}} _ {i } ^ { I } \right)\\
\mathcal { L } _ { S} \triangleq - \sum _ { j = 1 } ^ { n } \sum _ { i = 1 } ^ { n_{S} } { \hat { {\bf{y}} } _ {  j} ^ { i, S } \log \left( {\bf{o}} _ {  j} ^ { i,S } \right)}
\end{gather}
where ${\hat { {\bf{y}} } _ { i} ^ { I } }$ is the gold intent label, ${\hat { {\bf{y}} } _ { j} ^ { i, S} }$ is the gold slot label for $j$th token, $n_{I}$ is the number of intent labels, and $n_{S}$ is the number of slot labels.

The final training objective is as follow:
\begin{equation}
\label{loss}
    \mathcal{L}=\alpha \mathcal { L } _ { I} + \beta\mathcal { L } _ { S}+\lambda\mathcal { L } _ { intra}+\gamma\mathcal { L } _ { inter}
\end{equation}
where $\alpha$, $\beta$, $\lambda$, $\gamma$ are the hyper-parameters.
\section{Experiments}
\subsection{Datasets and Metrics}
The experiments in this study utilize the MultiATIS++ dataset, a benchmark dataset for cross-lingual SLU tasks. The dataset, available at \cite{xu2020end}, consists of 18 intents and 84 slots for each language. In addition to the existing Hindi (hi) and Turkish (tr) data in the Multilingual ATIS dataset, human-translated data for six languages (Spanish, German, Chinese, Japanese, Portuguese, and French) are included. In line with prior studies~\cite{qin2022gl,liang2022label,goo2018slot}, we employ accuracy to assess the performance of intent prediction, F1 score to evaluate slot filling, and overall accuracy to evaluate sentence-level semantic frame parsing. The overall accuracy metric reflects the correctness of all predictions, encompassing both intent and slots within the utterance.


\subsection{Implementation Details}
\begin{table*}[t]
    \centering
    \small
        \caption{Experiment results on the MultiATIS++ dataset. The results with ``$\diamondsuit$'' denotes that they are taken from the corresponding published paper, results with $^\dag$ are cited from \cite{qin2022gl}, and results with $^\ddag$ are cited from \cite{liang2022label}. `--' denotes missing results from the published work.} \label{table:result}
    \setlength{\tabcolsep}{0.45mm}{
    \begin{tabular}{l|ccccccccc|c}
    \toprule
    
\textbf{Intent~(Acc)} & en & de & es & fr & hi & ja & pt & tr & zh & \textbf{AVG} \\ 
    \midrule[0.5pt]
mBERT$^\dag$~\cite{devlin2019bert} & 98.54 & 95.40 & 96.30 & 94.31 & 82.41 & 76.18 & 94.95 & 75.10 & 82.53 & 88.42\\
    ZSJoint$^\ddag$~\cite{chen2019bert}      & 98.54 & 90.48 &  93.28 & 94.51 & 77.15 &76.59 & 94.62 & 73.29 & 84.55 & 87.00 \\
Ensemble-Net$\diamondsuit$~\cite{razumovskaia2022crossing} & 90.26 & 92.50 & 96.64 & 95.18 & 77.88 & 77.04 & 95.30 & 75.04 & 84.99 & 87.20 \\
CoSDA$^\dag$~\cite{ijcai2020p0533} & 95.74 & 94.06 & 92.29 & 77.04 & 82.75 & 73.25 & 93.05 & 80.42 & 78.95 & 87.32 \\
GL-CL{\scriptsize E}F$\diamondsuit$~\cite{qin2022gl} & 98.77 & 97.53 & 97.05 & 97.72 & 86.00 & 82.84 & 96.08 & 83.92 & 87.68 & 91.95 \\
    LAJ-MCL$\diamondsuit$~\cite{liang2022label}         & 98.77 & 98.10 & 98.10 & 98.77 & 84.54 & 81.86 & 97.09 & 85.45 & 89.03 &  92.41 \\
    \midrule[0.5pt]
I$^2$KD-SLU & \textbf{98.87} & \textbf{98.18} & \textbf{98.22} & \textbf{98.94} & \textbf{86.67} & \textbf{82.66} & \textbf{97.22} & \textbf{85.99} & \textbf{89.47} & \textbf{92.91} \\
    \midrule[0.75pt]
			\textbf{Slot~(F1)} & en & de & es & fr & hi & ja & pt & tr & zh & \textbf{AVG} \\     
\midrule[0.5pt]
Ensemble-Net$\diamondsuit$~\cite{razumovskaia2022crossing} & 85.05 & 82.75 & 77.56 & 76.19 & 14.14 & 9.44 & 74.00 & 45.63 & 37.29 & 55.78\\
mBERT$^\dag$~\cite{devlin2019bert} & 95.11 & 80.11 & 78.22 & 82.25 & 26.71 & 25.40 & 72.37 & 41.49 & 53.22 & 61.66 \\
    ZSJoint$^\ddag$~\cite{chen2019bert}       & 95.20 & 74.79 & 76.52 & 74.25 & 52.73 & 70.10 & 72.56 & 29.66 & 66.91 & 68.08 \\
CoSDA$^\dag$~\cite{ijcai2020p0533} & 92.29 & 81.37 & 76.94 & 79.36 & 64.06 & 66.62 & 75.05 & 48.77 & 77.32 & 73.47 \\
GL-CL{\scriptsize E}F$\diamondsuit$~\cite{qin2022gl}  &95.39 & 86.30 & 85.22 & 84.31 & 70.34 & 73.12 & 81.83 & 65.85 & 77.61 & 80.00 \\
LAJ-MCL$\diamondsuit$\cite{liang2022label}       & 96.02 & 86.59 & 83.03 & 82.11 & 61.04 & 68.52 & 81.49 & 65.20 & 82.00 & 78.23 \\
    \midrule[0.5pt]
I$^2$KD-SLU & \textbf{96.18} & \textbf{86.74} & \textbf{85.50} & \textbf{84.28} & \textbf{73.06} & \textbf{74.14} & \textbf{82.54} & \textbf{68.16} & \textbf{83.14} & \textbf{81.53} \\
    \midrule[0.75pt]
			\textbf{Overall~(Acc)} & en & de & es & fr & hi & ja & pt & tr & zh & \textbf{AVG} \\ 
    \midrule[0.5pt]
			AR-S2S-PTR$\diamondsuit$~\cite{rongali2020don} & 86.83 & 34.00 & 40.72 & 17.22 & 7.45 & 10.04 & 33.38 & -- & 23.74 & --\\
			IT-S2S-PTR$\diamondsuit$~\cite{zhu2020don} & 87.23 & 39.46 & 50.06 & 46.78 & 11.42 & 12.60 & 39.30 & -- & 28.72 & --\\
		mBERT$^\dag$~\cite{devlin2019bert} & 87.12 & 52.69 & 52.02 & 37.29 & 4.92 & 7.11 & 43.49 & 4.33 & 18.58 & 36.29 \\
    ZSJoint$^\ddag$~\cite{chen2019bert}      & 87.23 & 41.43 & 44.46 & 43.67 & 16.01 & 33.59 & 43.90 & 1.12 & 30.80  & 38.02 \\
    CoSDA$^\dag$~\cite{ijcai2020p0533} & 77.04 & 57.06 & 46.62 & 50.06 & 26.20 & 28.89 & 48.77 & 15.24 & 46.36 & 44.03 \\
    GL-CL{\scriptsize E}F$\diamondsuit$~\cite{qin2022gl}     & 88.02 & 66.03 & 59.53 & 57.02 & 34.83 & 41.42 & 60.43 & 28.95 & 50.62 & 54.09 \\
    LAJ-MCL$\diamondsuit$\cite{liang2022label}        & 89.81 & 67.75 &59.13& 57.56 & 23.29 & 29.34 & 61.93  & 28.95 & 54.76 & 52.50 \\
    \midrule[0.5pt]
I$^2$KD-SLU & \textbf{90.04} & \textbf{68.01} & \textbf{59.76} & \textbf{58.03} & \textbf{35.08} & \textbf{43.02} & \textbf{63.02} & \textbf{29.31} & \textbf{55.06} & \textbf{55.70} \\
    \bottomrule
    \end{tabular}}
    \label{tab:main_atis_mbert}
\end{table*}
The mBERT model in experiment has $N = 12$ attention heads and $M = 12$ transformer blocks. Following ~\cite{qin2022gl}, hyperparameters are selected by searching a combination of batch size, learning rate within the candidate set $\{2\times 10^{-7},5\times 10^{-7},  1\times 10^{-6},  2\times 10^{-6},5\times 10^{-6},6\times 10^{-6},5\times 10^{-5},5\times 10^{-4}\}$ and $\{4,8,16,32\}$. $\alpha$, $\beta$, $\lambda$, $\gamma$ are set to 0.9, 0.1, 0.7 and 0.3 in Eq.\ref{loss}, respectively.
To optimize the parameters in our model, we utilize the Adam optimizer~\cite{kingma2014adam} with $\beta_1=0.9$ and $\beta_2=0.98$.
The learning rate follows a schedule where it decreases proportionally to the inverse square root of the step number after the warm-up phase. In all experiments, we select the model that achieves the highest overall accuracy on the \texttt{dev} set and evaluate its performance on the \texttt{test} set.
All experiments are conducted at an Nvidia Tesla-V100 GPU. 


We compare our model to the following baselines:
\texttt{AR-S2S-PTR}
~\cite{rongali2020don},
\texttt{IT-S2S-PTR}
~\cite{zhu2020don},
\texttt{mBERT}~\cite{devlin2019bert},
\texttt{Ensemble-Net}~\cite{razumovskaia2022crossing},
\texttt{ZSJoint}~\cite{chen2019bert},
\texttt{CoSDA}~\cite{ijcai2020p0533},
\texttt{GL-CL{\scriptsize E}F}~\cite{qin2022gl},
\texttt{LAJ-MCL}~\cite{liang2022label}.
These methods are all commonly used approaches in the field of SLU in recent years. In this paper, we evaluate these methods on the MultiATIS++ dataset comparing with our proposed method.
\subsection{Main Results}
The performance comparison of I$^2$KD-SLU and baselines are shown in Table \ref{tab:main_atis_mbert}, from which we have the following observations: Despite the fact that the models which applies code-switching method outperform the models which do not use this method, I$^2$KD-SLU further improves the performance and obtains a relative improvement of 3.0$\%$ over the previous best model in terms of average overall accuracy. The potential reason is that our method enhance the mutual guidance between intent and slots by intra and inter knowledge distillation.
\subsection{Model Analysis}

\begin{table}[!t]
\centering
\caption{Ablation results on the MultiATIS++ dataset.}
    \setlength{\tabcolsep}{4.7mm}{
\begin{tabular}{lccc}
\toprule[1pt]
\textbf{Models} & Intent & Slot&Overall \\ \midrule[1pt]
\textbf{I$^2$KD-SLU} & \textbf{92.91}&\textbf{81.53} &\textbf{55.70}\\
\midrule
\textit{w/o intra KD} & 91.46($\downarrow$1.45)&78.14($\downarrow$3.39)&51.02($\downarrow$4.68)\\
\textit{w/o inter KD} & 91.53($\downarrow$0.38)&78.26($\downarrow$3.27)&51.72($\downarrow$3.98)\\
\textit{More Parameters} & 88.34($\downarrow$4.57)&74.84($\downarrow$6.69)&46.15($\downarrow$9.55)\\
\bottomrule[1pt]
\end{tabular}}
\label{tab:ablation}
\end{table}
\subsubsection{Effect of Intra Knowledge Distillation Module}
To demonstrate the effectiveness of intra knowledge distillation module, we remove it and refer it to \textit{w/o intra KD} in Table \ref{tab:ablation}. We can observe that after we remove the intra knowledge distillation module, the intent accuracy drops by 1.45$\%$ and the slot F1 drops by 3.39$\%$. Moreover, the overall accuracy also drops by 4.68$\%$. These results show the importance of intra knowledge distillation, which promotes knowledge transfer between different languages.
\subsubsection{Effect of Inter Knowledge Distillation Module}
To demonstrate the effectiveness of inter knowledge distillation, we remove it and refer it to \textit{w/o inter KD} and the results are shown in Table \ref{tab:ablation}. After we remove the inter knowledge distillation module, the intent accuracy of MixATIS++ dataset drops by 0.38$\%$ and the slot F1 of MixATIS++ dataset drops by 3.27$\%$. Moreover, the overall accuracy also drops by 3.98$\%$. We can clearly observe that inter knowledge distillation is beneficial in improving the performance of the model. By applying inter knowledge distillation, the model can predict the intent and slots more accurately, which achieves the mutual guidance between intent and slots.
\subsubsection{Effect of More Parameters}
To assess the impact of increased parameters in our method, we build upon previous research ~\cite{qin2020agif,qin2021gl} by introducing an additional LSTM layer after the final layer of mBERT, referred to as "More Parameters." The outcomes presented in Table \ref{tab:ablation} illustrate that our approach surpasses mBERT with more parameters in terms of intent accuracy, slot F1, and overall accuracy by 4.57\%, 6.69\%, and 9.55\%, respectively. These findings provide evidence that the enhancement achieved by our method can be attributed to both intra- and inter-knowledge distillation techniques.

\section{Further experiment}
\begin{figure*}[t]
  \centering
\includegraphics[width=\linewidth]{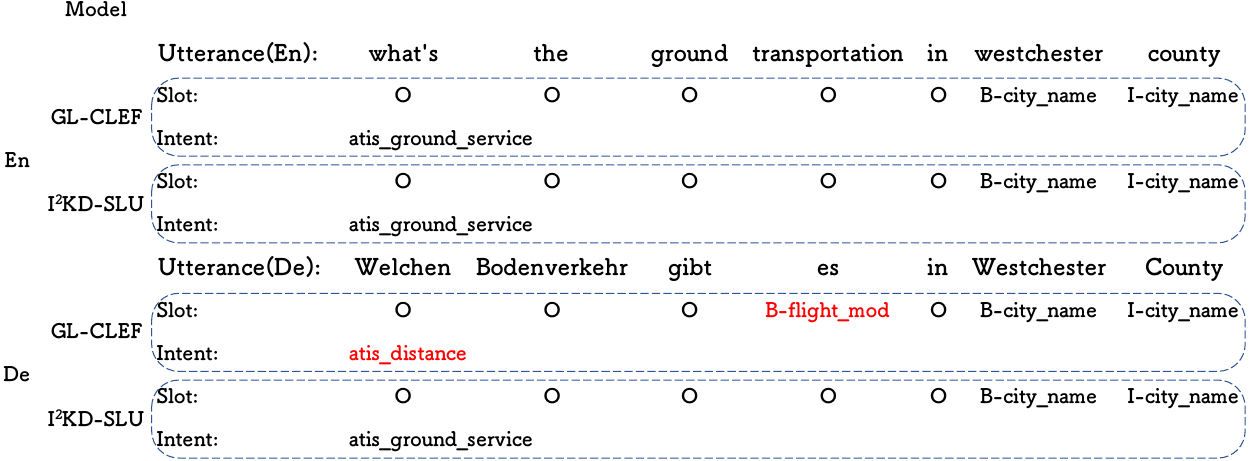}
  \caption{A case study of our framework and previous best model GL-CL{\scriptsize E}F. Intents and slots in \textcolor[RGB]{234, 51, 35}{red} are those that are predicted incorrectly.}
  \label{fig:case}
\end{figure*}

To further demonstrate the superiority of our framework, we present one case in Fig. \ref{fig:case}. We can clearly observe that both GL-CL{\scriptsize E}F and I$^2$KD-SLU predict the slots and intent correctly in English. However, GL-CL{\scriptsize E}F predicts the slots and intent incorrectly in German, while our model still predicts them correctly. The reason for this is that our model achieve the mutual guidance between intent and slots and promotes knowledge transfer at the same time.
\section{Conclusions}
In this paper, we propose a novel intra-inter knowledge distillation framework I$^2$KD-SLU for zero-shot cross-lingual spoken language understanding~(SLU), which achieves the mutual guidance between intent and slots and promotes the knowledge transfer between different languages. Experiments on MultiATIS++ dataset show that I$^2$KD-SLU achieve a new state-of-the-art performance. Model analysis demonstrates the superiority of I$^2$KD-SLU. In the future, we will explore the effectiveness of our method in other zero-shot cross-lingual tasks to further improve the performance.
\bibliographystyle{splncs04}
\bibliography{mybib}

\end{document}